\pdfoutput=1
\documentclass[11pt]{article}

\usepackage{acl}

\usepackage{times}
\usepackage{latexsym}

\usepackage[T1]{fontenc}
\usepackage[utf8]{inputenc}

\usepackage{microtype}

\usepackage{inconsolata}
\usepackage{hyperref}
\usepackage[ruled, lined, longend]{algorithm2e}
\usepackage{amsfonts} 
\usepackage{multirow}
\usepackage{pifont}
\usepackage{booktabs}
\usepackage{amsmath}
\usepackage{xcolor} 
\usepackage{tikz}
\usetikzlibrary{shapes,shadows}
\usepackage[most]{tcolorbox}

\title{ICDPO: Effectively Borrowing Alignment Capability of Others\\via In-context Direct Preference Optimization}

\author{
    Feifan Song$^{1,2}$, Yuxuan Fan$^{1,2}$, Xin Zhang$^{3}$, Peiyi Wang$^{1,2}$, \textbf{Houfeng Wang}$^{1,2}$\thanks{\quad Corresponding author.} \\
  $^{1}$National Key Laboratory of Multimedia Information Processing, Peking University \\
  $^{2}$School of Computer Science, Peking University $^{3}$Microsoft Research Asia\\
  \texttt{\{songff,yxfan\}@stu.pku.edu.cn; xinzhang3@microsoft.com} \\
  \texttt{wangpeiyi9979@gmail.com; wanghf@pku.edu.cn}}
\newcommand\modelname{ICDPO}

\begin{document}
\maketitle
\begin{abstract}
Large Language Models~(LLMs) rely on Human Preference Alignment~(HPA) to ensure the generation of safe content. 
Due to the heavy cost associated with fine-tuning, fine-tuning-free methods have emerged, typically modifying LLM decoding with external auxiliary methods.
However, these methods do not essentially enhance the LLM itself.
In this paper, we rethink the derivation procedures of DPO, based on which we conversely build an instant scorer using the states of the LLM before and after In-context Learning~(ICL).
Accordingly, we propose a novel approach called In-Context Direct Preference Optimization~(\modelname{}). It enables LLMs to borrow the HPA capabilities from superior LLMs with ICL, generating well-aligned responses as estimated by the aforementioned instant scorer, thereby enhancing the final performance.
\modelname{} can be further enhanced with a two-stage retriever and an upgraded scorer, both offering benefits.
Extensive experiments show its effectiveness, particularly in outperforming two fine-tuning-free baselines, and it exhibits competitiveness with SFT + LoRA.
We also conduct detailed analyses to offer comprehensive insights into \modelname{}.\footnote{The code of this work is available at \url{https://github.com/F2-Song/ICDPO}.}

\end{abstract}
\section{Introduction}
Human Preference Alignment~(HPA) is crucial within the LLM industry as it prevents LLMs from generating offensive, harmful, or misleading content contrary to human values.
Presently, mainstream approaches to HPA heavily depend on fine-tuning, exemplified by RLHF~\cite{stiennon2020learning, ouyang2022training, zhu2023principled}, RAFT~\cite{dong2023raft}, RRHF~\cite{yuan2023rrhf}, or DPO~\cite{rafailov2023direct}.
Nevertheless, the huge computational and data annotation costs associated with fine-tuning are hard to ignore.

As a response, fine-tuning-free approaches have gained popularity.
\citet{li2023rain} enable the LLM to take self-evaluation in decoding process. Alternatively, LLMs can borrow the capabilities of superior models~(i.e. teacher models) to improve responses. Here the concept of \textit{borrowing} is different from \textit{learning} for it does not bring real parameter updates. For instance, external scorers capable of distinguishing human preference can be involved to apply best-of-N selection for multiple candidates or enhance block selection during LLM inference~\cite{mudgal2023controlled}.

\begin{figure*}[htbp]
    \centering
    \includegraphics[width=0.95\textwidth]{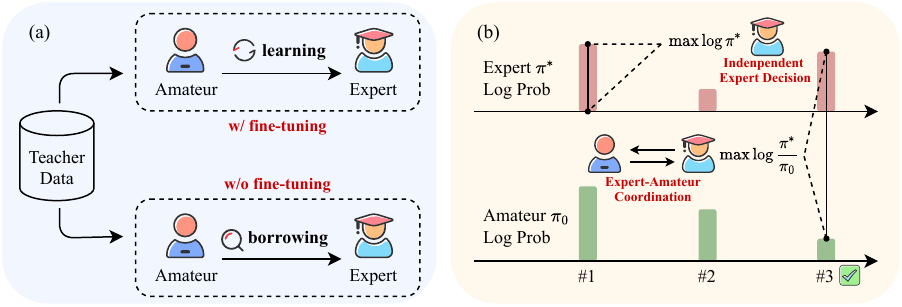} 
    \caption{The overview of \modelname{}. (a)~The difference in teacher data utilization between normal fine-tuning and ICL without fine-tuning. (b)~The core of \modelname{} is that expert-amateur coordination maximizes $S$ which represents the disparity between the expert and the amateur. It brings more accurate estimation than using only the expert LLM.}
    \label{fig:intro}
\end{figure*}

However, these approaches concentrate on the decoding stage, neglecting to fundamentally enhance the HPA capabilities of the LLM itself.
This limitation raises the question: \textbf{Can LLMs borrow the HPA capabilities of superior LLMs to develop themselves without fine-tuning?}
Therefore, we select In-context Learning~(ICL) to reach the target of \textit{borrowing}, as depicted in Figure~\ref{fig:intro}(a).
Unlike \textit{learning}, ICL enables LLMs to ingest well-aligned samples from external teachers, mimicking them to produce aligned responses without fine-tuning.

More importantly, we rethink the procedures of Direct Preference Optimization~(DPO) proposed in~\citet{rafailov2023direct}.
It integrates the policy LLM into the Reward Modeling by transforming RLHF objectives, bridging the relation between the provided reward model~(RM) and optimal policy $\pi^*$. Here, the RM quantifies the distributional disparity between $\pi^*$ and its reference model $\pi_0$.
Conversely, an optimized policy that aligns with human preference can collaborate with its pre-optimized reference model, potentially offering more reliable estimations of HPA for candidate responses.

Additionally, LLMs essentially undergo instantaneous meta-optimization via ICL, involving an internal parameter updating formulation similar to real fine-tuning~\cite{dai-etal-2023-gpt}.
Consequently, the states of an LLM before and after ICL can be regarded as the \textbf{Expert} $\pi^*$ and \textbf{Amateur} $\pi_0$, respectively, to form a customized RM for scoring multiple samples (named Contrastive Score $S$), thereby maximizing the effectiveness of ICL, as illustrated in Figure~\ref{fig:intro}(b).
This process remains fine-tuning-free and entails only one LLM during decoding, which we term as \textbf{I}n-\textbf{C}ontext \textbf{D}irect \textbf{P}references \textbf{O}ptimization~(\textbf{\modelname{}}).

Since we intend to harness the LLM through contextual demonstrations, the selection and ordering of demonstrated samples become crucial.
Inspired by the nature of fine-tuning, where aligned distributions between training and test sets maximize effectiveness, we develop a two-stage retriever to identify demonstrations that are most similar to the test samples in both form and semantics, thereby improving the performance of \modelname{}. 
Furthermore, like the prevalent contrastive fine-tuning in HPA, we elevate $S$ to $\hat{S}$ by incorporating both favorable and unfavorable samples to amplify the disparities between $\pi^*$ and $\pi_0$. 
It works as debiasing the distribution of candidates to further enhance \modelname{}.

Extensive experiments are conducted to evaluate the proposed \modelname{}, encompassing evaluations using both RM and GPT-4, along with an ablation study to validate each module.
We also provide comprehensive analyses of multiple aspects in \modelname{}.
The main observations are as follows:\\
(1)~\modelname{} borrows the HPA ability from superior LLMs through ICL, which in turn produces the $\pi^*$ collaborating with the initial $\pi_0$ to conduct scoring.
This significantly enhances performance by improving and exploiting the LLM itself, surpassing two fine-tuning-free baselines, as well as being competitive with SFT plus LoRA\cite{hu2021lora}.\\
(2)~Contextual demonstrations are closely related to the final performance. Specifically, demonstrated samples of higher quality and the proposed two-stage retriever can both facilitate \modelname{}.\\
(3)~Regarding scoring, the scorers $S$ and $\hat{S}$ in \modelname{} can provide reliable estimations of the degree of HPA, which can also be applied to fine-tuning methods, like DPO.
\section{Methodology}
In this section, we rethink the transformation from RLHF to DPO~\cite{rafailov2023direct}, an elegant supervised fine-tuning algorithm derived from the original RLHF objective $\mathcal{T}$.
We focus on the relation between a given RM and the corresponding optimal policy $\pi^*$, and adapt it to LLM inference in the manner of In-context Learning~(ICL), which we term as \modelname{}.

\subsection{From Reward Model to Policy LLM}
The original target $\mathcal{T}$ of RLHF is to optimize the policy LLM $\pi$ for the acquisition of a synthetic reward $\mathcal{R}$, the combination of a fundamental reward from the given RM $r^*$ and a KL-regularization to reference policy $\pi_0$,
\begin{equation}
\label{eq:rlhf}
\begin{aligned}
    \mathcal{T} &= \max_{\pi} \mathbb{E}[\mathcal{R}]\\
    &= \max_{\pi} \mathbb{E}[r^*(x, y) - \beta \log\frac{\pi(y\mid x)}{\pi_0(y\mid x)}]
\end{aligned}
\end{equation}
\citet{rafailov2023direct} construct the Direct Preference Optimization~(DPO) algorithm by first transforming Equation~\ref{eq:rlhf},
\begin{equation}
\label{eq:dpo}
\begin{aligned}
\mathcal{T} &=\min_{\pi} \mathbb{E}[\log\frac{\pi(y\mid x)}{\pi_0(y\mid x)} - \frac{1}{\beta}r^*(x, y)]\\
&=\min_{\pi} \mathbb{E}
[\log\frac{\pi(y\mid x)Z(x)}{\pi_0(y\mid x)\exp\left(\frac{1}{\beta}r^*(x, y)\right)} \\
&\quad- \log Z(x) ]
\end{aligned}
\end{equation}
where
\begin{equation}
    Z(x) = \sum_y \pi_0(y\mid x)\exp\left(\frac{1}{\beta}r^*(x, y)\right)
\end{equation}
is the partition function, and the relation between $r^*$ and 
the optimal policy $\pi^*$ of Equation~\ref{eq:dpo} is found:
\begin{equation}
\label{eq:relation}
    r^*(x, y) = \beta \log \frac{\pi^*(y \mid x)}{\pi_0(y \mid x)} + \beta \log Z(x)
\end{equation}

\subsection{Preference Optimization via ICL}
In RLHF, $r^*$ typically represents the outcome of Reward Modeling preceding the PPO stage, and $\pi^*$ denotes the corresponding optimal policy.
DPO opts to integrate $\pi$ into the supervised objective of Reward Modeling and devises an SFT-style fine-tuning approach based on the formulation of Equation~\ref{eq:relation}.
Conversely, we rethink Equation~\ref{eq:rlhf} and \ref{eq:relation} with the aim of avoiding parameter modification in the policy LLM $\pi$.

With an optimized policy LLM $\pi^*$ and a reference policy $\pi_0$, according to Equation~\ref{eq:relation}, we can build a customized reward function $\hat{r}$ as follows:
\begin{equation}
\label{eq:new_rm}
    \hat{r}(x,y) = \log \frac{\pi^*(y \mid x)}{\pi_0(y \mid x)} + \log Z(x)
\end{equation}
Since $\pi^*$ has been optimized to align with human preference, the corresponding $\hat{r}$ should well reflect the extent of human preference to some degree.
Additionally, the synthetic $\mathcal{R}$ in Equation~\ref{eq:rlhf} incorporates the KL-regularization component to prevent the policy from deviating too far from the typical linguistic space.
Therefore, if $\pi^*$ is presumed to retain this capability without the concern for regularization, Equation~\ref{eq:rlhf} could exclusively concentrate on preference rewards. Consequently, with Equation~\ref{eq:new_rm}, we could have
\begin{equation}
\label{eq:icdpo}
\begin{aligned}
    \max_{y} \mathcal{R} &\equiv \max_{y} \hat{r}(x,y)\\
    &\equiv \max_{y} \log \frac{\pi^*(y \mid x)}{\pi_0(y \mid x)}
\end{aligned}
\end{equation}
because $Z(x)$ in Equation~\ref{eq:new_rm} involves only $x$.

Furthermore, $\pi^*$ ought to be optimized while the initial objective necessitates it not to be fine-tuned. We thus use ICL to fulfill all these criteria, with inspiration from~\citet{dai-etal-2023-gpt} that inner meta-optimization can be demonstrated in ICL with contextual demonstrations $\textbf{d}$ and tested $x$:\\
\begin{equation}
\begin{aligned}
&\text{Attention}([\textbf{d};x], q) \\
&\approx W_V[\textbf{d};x](W_K[\textbf{d};x])^Tq \\
&=\left(W_Vx(W_Kx)^T + W_V\textbf{d}(W_K\textbf{d})^T\right)q\\
&=\left(W_\text{ZSL}+ \Delta W_\text{ICL}\right)q
\end{aligned}
\end{equation}
Here, $q=W_Qt$ represents the query of the next token $t$ in the self-attention mechanism, and $W_\text{ZSL}q = W_Vx(W_Kx)^Tq$ approximates the attention result in a zero-shot setting (i.e., no demonstrations involved). Furthermore, $\Delta W_\text{ICL} = W_V\textbf{d}(W_K\textbf{d})^T$ updates the weights of $W_\text{ZSL}$ using demonstrations $\textbf{d}$ in the context, thereby facilitating meta-optimization.

As a result, the optimized $\pi^*$ can be built directly through ICL, while the reference LLM $\pi_0$ serves as the initial checkpoint, i.e., the base model in this scenario.
Moreover, $\pi^*$ does not undergo parameter updates from fine-tuning, thereby preserving the initial language modeling capacity as $\pi_0$, without the need for additional regularization.
Therefore, we can employ a two-stage inference pipeline. In the first stage, multiple responses $\textbf{y}$ are sampled from $\pi^*$ as candidates to guarantee a potentially acceptable output, termed as \textbf{Generation}. Subsequently, in the second \textbf{Scoring} stage, the contrastive score $S$ for each candidate $y \in \textbf{y}$ is computed based on the demonstrated samples $\textbf{d}$, the prompt $x$, and Equation~\ref{eq:icdpo}:
\begin{equation}
\begin{aligned}
\label{eq:label}
    S(\textbf{d},x,y) &= \log \frac{\pi^*(y \mid x)}{\pi_0(y \mid x)}\\
    &=\log \frac{\pi(y \mid [\textbf{d}; x])}{\pi(y \mid x)}
\end{aligned}
\end{equation}
wherein the most preferred response $y^*$ can be chosen based on the largest $S$, indicating the highest reward of human preference, as in Figure~\ref{fig:intro}(b). We summarize the entire workflow as \textbf{\modelname{}}.
Note that $\pi^*$ is acquired through ICL, implying that only a single checkpoint is required throughout the entire inference process.
We define the score of response $y$ towards prompt $x$ from $\pi$ as its probability of generating $y$,
\begin{equation}
    \pi(y\mid x) = \sum_{i} P_{\pi}(y_i|x,y_{<i})
\end{equation}
\subsection{Connection to Contrastive Decoding}
We observe that Equation~\ref{eq:icdpo} relies on a contrastive estimation involving two LLMs: $\pi^*$ and $\pi_0$.
Furthermore, \citet{li-etal-2023-contrastive} enhance the quality of generated texts by replacing the naive maximum probability decoding with a contrastive objective, namely Contrastive Decoding~(CD), where each step utilizes both an expert model $\pi^{+}$ and an amateur model $\pi^{-}$,
\begin{equation}
\label{eq:cd}
    y^*_i = \arg\max_{y_i} \log \frac{\pi^{+}(y_i \mid x, y_{<i})}{\pi^{-}(y_i \mid x, y_{<i})}
\end{equation}
\begin{algorithm}
\caption{\modelname{}}
\label{algorithm:icdpo}
\LinesNumbered
\KwIn{
    Language Model~$\pi$, 
    Dataset~$D$, 
    input prompt~$x$
}
\KwOut{
    Response~$y$ with the largest score
}
\tcp{Generation stage}
Retrieve $m$ demonstrated samples $\textbf{d}$ from $D$

Sample $n$ responses $\{y_i\}$ from $\pi(y \mid [\textbf{d}; x])$

\tcp{Scoring stage}
Let $s=-\infty$

Let $p=0$

\For {$y_i \in \left \{y_1, ..., y_{n}\right \}$}{
    Estimate $\pi(y \mid [\textbf{d}; x])$ in ICL

    Estimate $\pi(y \mid x)$
    
    Estimate $S(\textbf{d},x,y)$ with Equation~\ref{eq:label}
    
    \If{$S(\textbf{d},x,y)>s$}{
        $s=S(\textbf{d},x,y)$
        
        $p=i$
    }
}
Let $y = y_p$

\textbf{return} $y$

\end{algorithm}
\\
While Equation~\ref{eq:icdpo} optimizes at the sentence-level instead of estimating token-wise scores as in CD for the generated $y$, we note that $\pi^*$ and $\pi_0$ are essentially treated as the expert and amateur models, respectively, in terms of HPA. This enhances LLM decoding with a focus on human preference.
To achieve this, we can enhance Equation~\ref{eq:icdpo} and Equation~\ref{eq:label} by introducing a purposely worse policy $\pi^-$ for HPA to replace the original $\pi_0$.
More precisely, $\pi^-$ can also be acquired through In-context Learning with human-rejected samples $\textbf{d}^-$ as demonstrations, whereas the original expert model $\pi^*$ in Equation~\ref{eq:icdpo} can be relabeled as $\pi^+$ and its contextual demonstrations comprise solely human-chosen $\textbf{d}^+$.
Hence, the promoted contrastive score is
\begin{equation}
\begin{aligned}
    \hat{S}(\textbf{d}^+,\textbf{d}^-,x,y) &= \log \frac{\pi^+(y \mid x)}{\pi^-(y \mid x)}\\
    &=\log \frac{\pi(y \mid [\textbf{d}^+; x])}{\pi(y \mid [\textbf{d}^-; x])}
\end{aligned}
\end{equation}

\subsection{Retrieval}
The demonstrated samples and their sequencing are acknowledged as crucial factors for ICL.
Since the process of ICL may resemble gradient descent during actual model training, we can further amplify the inner meta-optimization from the fine-tuning standpoint.
Given that the closeness between the distributions of the test data and the training data is vital for the efficacy of fine-tuning, it should coherently work in ICL.
Consequently, we also employ a prevalent similarity-based retriever to determine the sample selection and their corresponding sequencing, while incorporating additional considerations:
(1) Despite their effectiveness, pre-trained retrievers (e.g., SBERT-based methods) have significant computational costs for the large number of samples, requiring a two-stage design where coarse-grained selections are first made before more fine-grained retrievals.
(2) Since LLMs operate in an auto-regressive manner, the last portion of the tested samples should have the most significant impact.
Hence, retrieving those with structurally similar end portions is prioritized, and able to additionally reduce computational overhead.

Therefore, we propose a two-stage retriever containing a coarse-grained BM25 retriever~\cite{robertson2009probabilistic} focusing on the end of each sample, and an SBERT~\cite{reimers-gurevych-2019-sentence} to execute fine-grained retrieval:
\begin{equation}
\begin{aligned}
    R(\{x_i\}) &= \text{SBERT}(\{a_j\})\\
    \{a_j\} &= \text{BM25}(\{x_i[-L:]\})
\end{aligned}
\end{equation}
where $\{x_i\}$ is the support set, and $L$ is the window size constraining the ending range of samples for BM25. We show that \modelname{} equipped with $\mathcal{R}$ yields notable improvement overall.
\begin{table}
\centering
\scalebox{0.9}{
  \begin{tabular}{llll}
    \toprule
    \multicolumn{1}{l}{\textbf{Method}} & \multicolumn{1}{l}{\textbf{Harmless}} & \multicolumn{1}{c}{\textbf{Helpful}} & \multicolumn{1}{c}{\textbf{Total}} \\
    \midrule
    \multicolumn{4}{c}{w/ \textit{HH-RLHF}$_\text{raw}$}\\
    \midrule
    \multicolumn{1}{l}{Raw} & \multicolumn{1}{c}{24.23} & \multicolumn{1}{c}{-47.62} & \multicolumn{1}{c}{-11.70}\\
    \multicolumn{1}{l}{LLaMA2-chat} & \multicolumn{1}{c}{105.97} & \multicolumn{1}{c}{61.18} & \multicolumn{1}{c}{83.57}\\
    \multicolumn{1}{l}{GPT-3.5-turbo} & \multicolumn{1}{c}{105.99} & \multicolumn{1}{c}{73.80} & \multicolumn{1}{c}{89.89}\\
    \midrule
    \multicolumn{4}{c}{w/ \textit{SytheticGPT}$_\text{raw}$}\\
    \midrule
    \multicolumn{1}{l}{Raw} & \multicolumn{1}{c}{-} & \multicolumn{1}{c}{-} & \multicolumn{1}{c}{74.04}\\
    \multicolumn{1}{l}{LLaMA2-chat} & \multicolumn{1}{c}{-} & \multicolumn{1}{c}{-} & \multicolumn{1}{c}{120.31}\\
    \bottomrule
  \end{tabular}
}
\caption{\label{main_results2}
    Reference results.
}
\end{table}
\begin{table*}[t]
\centering
\scalebox{0.9}{
  \begin{tabular}{llllllllll}
    \toprule
    \multicolumn{1}{l}{\multirow{2}{*}{\textbf{Method}}} & \multicolumn{3}{c}{\textbf{LLaMA}} & \multicolumn{3}{c}{\textbf{LLaMA2}} & \multicolumn{3}{c}{\textbf{Mistral}}\\
    \cmidrule(lr){2-4}\cmidrule(lr){5-7}\cmidrule(lr){8-10}
    \multicolumn{1}{l}{} & \multicolumn{1}{c}{\textbf{Harmless}} & \multicolumn{1}{c}{\textbf{Helpful}} & \multicolumn{1}{c}{\textbf{Total}} & \multicolumn{1}{c}{\textbf{Harmless}} & \multicolumn{1}{c}{\textbf{Helpful}} & \multicolumn{1}{c}{\textbf{Total}} & \multicolumn{1}{c}{\textbf{Harmless}} & \multicolumn{1}{c}{\textbf{Helpful}} & \multicolumn{1}{c}{\textbf{Total}}\\
    \midrule
    \multicolumn{10}{c}{w/ \textit{HH-RLHF}$_\text{raw}$}\\
    \midrule
    \multicolumn{1}{l}{Base} & \multicolumn{1}{c}{4.47} & \multicolumn{1}{c}{-77.53} & \multicolumn{1}{c}{-36.54} & \multicolumn{1}{c}{6.25} & \multicolumn{1}{c}{-67.67} & \multicolumn{1}{c}{-30.72} & \multicolumn{1}{c}{9.59} & \multicolumn{1}{c}{-33.22} & \multicolumn{1}{c}{-11.82}\\
    \multicolumn{1}{l}{SFT} & \multicolumn{1}{c}{20.23} & \multicolumn{1}{c}{-65.46} & \multicolumn{1}{c}{-22.63} & \multicolumn{1}{c}{20.48} & \multicolumn{1}{c}{-60.77} & \multicolumn{1}{c}{-20.16} & \multicolumn{1}{c}{21.91} & \multicolumn{1}{c}{-48.65} & \multicolumn{1}{c}{-13.38}\\
    \multicolumn{1}{l}{ICDPO} & \multicolumn{1}{c}{25.02} & \multicolumn{1}{c}{-64.95} & \multicolumn{1}{c}{-19.97} & \multicolumn{1}{c}{39.81} & \multicolumn{1}{c}{-71.89} & \multicolumn{1}{c}{-16.05} & \multicolumn{1}{c}{26.60} & \multicolumn{1}{c}{-51.38} & \multicolumn{1}{c}{-12.40}\\
    \multicolumn{1}{l}{ICDPO$+{\hat{S}}$} & \multicolumn{1}{c}{24.03} & \multicolumn{1}{c}{-55.86} & \multicolumn{1}{c}{-15.92} & \multicolumn{1}{c}{42.52} & \multicolumn{1}{c}{-63.53} & \multicolumn{1}{c}{-10.52} & \multicolumn{1}{c}{32.78} & \multicolumn{1}{c}{-42.82} & \multicolumn{1}{c}{-5.03}\\
    \multicolumn{1}{l}{ICDPO$+{\hat{S}}R$} & \multicolumn{1}{c}{22.50} & \multicolumn{1}{c}{-55.77} & \multicolumn{1}{c}{-16.64} & \multicolumn{1}{c}{31.54} & \multicolumn{1}{c}{-63.22} & \multicolumn{1}{c}{-15.85} & \multicolumn{1}{c}{25.15} & \multicolumn{1}{c}{-44.75} & \multicolumn{1}{c}{-9.81}\\
    \midrule
    \multicolumn{10}{c}{w/ LLaMA2-chat}\\
    \midrule
    \multicolumn{1}{l}{SFT} & \multicolumn{1}{c}{48.92} & \multicolumn{1}{c}{20.54} & \multicolumn{1}{c}{34.73} & \multicolumn{1}{c}{72.94} & \multicolumn{1}{c}{42.24} & \multicolumn{1}{c}{57.59} & \multicolumn{1}{c}{77.59} & \multicolumn{1}{c}{49.29} & \multicolumn{1}{c}{63.43}\\
    \multicolumn{1}{l}{RM-Aug} & \multicolumn{1}{c}{5.06} & \multicolumn{1}{c}{-60.35} & \multicolumn{1}{c}{-27.66} & \multicolumn{1}{c}{2.92} & \multicolumn{1}{c}{-52.12} & \multicolumn{1}{c}{-24.61} & \multicolumn{1}{c}{13.65} & \multicolumn{1}{c}{-7.00} & \multicolumn{1}{c}{3.32}\\
    \multicolumn{1}{l}{RM-BoN} & \multicolumn{1}{c}{-1.47} & \multicolumn{1}{c}{-60.60} & \multicolumn{1}{c}{-31.04} & \multicolumn{1}{c}{2.90} & \multicolumn{1}{c}{-48.53} & \multicolumn{1}{c}{-22.82} & \multicolumn{1}{c}{7.16} & \multicolumn{1}{c}{-6.11} & \multicolumn{1}{c}{0.52}\\
    \multicolumn{1}{l}{ICDPO} & \multicolumn{1}{c}{68.75} & \multicolumn{1}{c}{-17.61} & \multicolumn{1}{c}{25.56} & \multicolumn{1}{c}{97.06} & \multicolumn{1}{c}{27.49} & \multicolumn{1}{c}{62.27} & \multicolumn{1}{c}{99.29} & \multicolumn{1}{c}{38.34} & \multicolumn{1}{c}{68.81}\\
    \multicolumn{1}{l}{ICDPO$+{\hat{S}}$} & \multicolumn{1}{c}{68.73} & \multicolumn{1}{c}{-11.75} & \multicolumn{1}{c}{28.48} & \multicolumn{1}{c}{98.03} & \multicolumn{1}{c}{29.36} & \multicolumn{1}{c}{63.69} & \multicolumn{1}{c}{97.26} & \multicolumn{1}{c}{45.08} & \multicolumn{1}{c}{71.16}\\
    \multicolumn{1}{l}{ICDPO$+{\hat{S}}R$} & \multicolumn{1}{c}{90.54} & \multicolumn{1}{c}{12.59} & \multicolumn{1}{c}{51.56} & \multicolumn{1}{c}{101.08} & \multicolumn{1}{c}{38.26} & \multicolumn{1}{c}{69.66} & \multicolumn{1}{c}{101.68} & \multicolumn{1}{c}{45.51} & \multicolumn{1}{c}{73.59}\\
    \midrule
    \multicolumn{10}{c}{w/ GPT-3.5-turbo}\\
    \midrule
    \multicolumn{1}{l}{SFT} & \multicolumn{1}{c}{54.28} & \multicolumn{1}{c}{-16.17} & \multicolumn{1}{c}{19.05} & \multicolumn{1}{c}{72.72} & \multicolumn{1}{c}{33.03} & \multicolumn{1}{c}{52.87} & \multicolumn{1}{c}{90.98} & \multicolumn{1}{c}{62.00} & \multicolumn{1}{c}{76.49}\\
    \multicolumn{1}{l}{ICDPO} & \multicolumn{1}{c}{63.91} & \multicolumn{1}{c}{-23.27} & \multicolumn{1}{c}{20.31} & \multicolumn{1}{c}{91.56} & \multicolumn{1}{c}{16.33} & \multicolumn{1}{c}{53.94} & \multicolumn{1}{c}{85.10} & \multicolumn{1}{c}{21.23} & \multicolumn{1}{c}{53.16}\\
    \multicolumn{1}{l}{ICDPO$+{\hat{S}}$} & \multicolumn{1}{c}{64.03} & \multicolumn{1}{c}{-14.86} & \multicolumn{1}{c}{24.58} & \multicolumn{1}{c}{92.14} & \multicolumn{1}{c}{21.40} & \multicolumn{1}{c}{56.76} & \multicolumn{1}{c}{85.83} & \multicolumn{1}{c}{36.14} & \multicolumn{1}{c}{60.98}\\
    \multicolumn{1}{l}{ICDPO$+{\hat{S}}R$} & \multicolumn{1}{c}{82.21} & \multicolumn{1}{c}{3.63} & \multicolumn{1}{c}{42.91} & \multicolumn{1}{c}{98.77} & \multicolumn{1}{c}{28.08} & \multicolumn{1}{c}{63.42} & \multicolumn{1}{c}{92.21} & \multicolumn{1}{c}{39.55} & \multicolumn{1}{c}{65.88}\\
    \bottomrule
  \end{tabular}
}
\caption{\label{main_results1}
    Main results on \textit{HH-RLHF} scored by RM$_\text{test}$. \textbf{Higher} values represent \textbf{better} performance towards HPA. 
}
\end{table*}
\section{Experiment}
\subsection{Settings} 
We employ two datasets, \textit{HH-RLHF} and \textit{SyntheticGPT} to comprehensively assess the effectiveness of \modelname{}.
Regarding the superior teacher models, we included LLaMA2-7B-chat~(denoted as \textbf{LLaMA2-chat}) and \textbf{GPT-3.5-turbo} to support all methods with base models.
For \textit{HH-RLHF}, we present the original version (referred to as \textit{HH-RLHF}$\text{raw}$) and its enhanced version from LLaMA2-chat and GPT-3.5-turbo, while for \textit{SyntheticGPT}, we consider both the original version (referred to as \textit{SyntheticGPT}$\text{raw}$) and the version adapted from LLaMA2-chat.

We implement three base models for comprehensive evaluation: LLaMA-7B~\cite{touvron2023llama}, LLaMA-2-7B~\cite{touvron2023llama2}, and Mistral-7B-v0.1~\cite{jiang2023mistral}, which we label as \textbf{LLaMA}, \textbf{LLaMA2}, and \textbf{Mistral}, respectively.
The details of data preparation and implementation (including the reward model RM$_\text{test}$) can be found in Appendix~\ref{appendix:data} and \ref{appendix:implementation}, respectively.
We evaluate the performance of both the original candidates and new ones from teacher models in these datasets using RM$_{\text{test}}$, as presented in Table~\ref{main_results2}.

\begin{table*}
\centering
\scalebox{0.88}{
  \begin{tabular}{llllllllll}
    \toprule
    \multicolumn{1}{l}{\multirow{2}{*}{\textbf{Method}}} & \multicolumn{3}{c}{\textbf{LLaMA}} & \multicolumn{3}{c}{\textbf{LLaMA2}} & \multicolumn{3}{c}{\textbf{Mistral}}\\
    \cmidrule(lr){2-4}\cmidrule(lr){5-7}\cmidrule(lr){8-10}
    \multicolumn{1}{l}{} & \multicolumn{1}{c}{\textbf{Harmless}} & \multicolumn{1}{c}{\textbf{Helpful}} & \multicolumn{1}{c}{\textbf{Total}} & \multicolumn{1}{c}{\textbf{Harmless}} & \multicolumn{1}{c}{\textbf{Helpful}} & \multicolumn{1}{c}{\textbf{Total}} & \multicolumn{1}{c}{\textbf{Harmless}} & \multicolumn{1}{c}{\textbf{Helpful}} & \multicolumn{1}{c}{\textbf{Total}}\\
    \midrule
    \multicolumn{10}{c}{w/ LLaMA2-chat}\\
    \midrule
    \multicolumn{1}{l}{ICDPO$+R$} & \multicolumn{1}{c}{90.55} & \multicolumn{1}{c}{9.96} & \multicolumn{1}{c}{50.24} & \multicolumn{1}{c}{100.62} & \multicolumn{1}{c}{35.89} & \multicolumn{1}{c}{68.25} & \multicolumn{1}{c}{101.49} & \multicolumn{1}{c}{40.34} & \multicolumn{1}{c}{70.91}\\
    \multicolumn{1}{l}{ICDPO$+\text{BM25}$} & \multicolumn{1}{c}{84.99} & \multicolumn{1}{c}{3.18} & \multicolumn{1}{c}{44.08} & \multicolumn{1}{c}{99.78} & \multicolumn{1}{c}{31.89} & \multicolumn{1}{c}{65.83} & \multicolumn{1}{c}{102.54} & \multicolumn{1}{c}{43.74} & \multicolumn{1}{c}{73.13}\\
    \multicolumn{1}{l}{ICDPO} & \multicolumn{1}{c}{68.75} & \multicolumn{1}{c}{-17.61} & \multicolumn{1}{c}{25.56} & \multicolumn{1}{c}{97.06} & \multicolumn{1}{c}{27.49} & \multicolumn{1}{c}{62.27} & \multicolumn{1}{c}{99.29} & \multicolumn{1}{c}{38.34} & \multicolumn{1}{c}{68.81}\\
    \multicolumn{1}{l}{ICL} & \multicolumn{1}{c}{62.30} & \multicolumn{1}{c}{-26.09} & \multicolumn{1}{c}{18.09} & \multicolumn{1}{c}{97.23} & \multicolumn{1}{c}{16.72} & \multicolumn{1}{c}{56.97} & \multicolumn{1}{c}{94.79} & \multicolumn{1}{c}{32.68} & \multicolumn{1}{c}{63.73}\\
    \multicolumn{1}{l}{ICL$_\text{uni}$} & \multicolumn{1}{c}{63.04} & \multicolumn{1}{c}{-25.25} & \multicolumn{1}{c}{18.89} & \multicolumn{1}{c}{95.64} & \multicolumn{1}{c}{14.74} & \multicolumn{1}{c}{55.18} & \multicolumn{1}{c}{94.54} & \multicolumn{1}{c}{33.06} & \multicolumn{1}{c}{63.80}\\
    \midrule
    \multicolumn{10}{c}{w/ GPT-3.5-turbo}\\
    \midrule
    \multicolumn{1}{l}{ICDPO$+R$} & \multicolumn{1}{c}{80.64} & \multicolumn{1}{c}{-1.13} & \multicolumn{1}{c}{39.75} & \multicolumn{1}{c}{98.08} & \multicolumn{1}{c}{24.45} & \multicolumn{1}{c}{61.25} & \multicolumn{1}{c}{89.91} & \multicolumn{1}{c}{31.10} & \multicolumn{1}{c}{60.50}\\
    \multicolumn{1}{l}{ICDPO$+\text{BM25}$} & \multicolumn{1}{c}{74.28} & \multicolumn{1}{c}{-3.24} & \multicolumn{1}{c}{35.51} & \multicolumn{1}{c}{96.18} & \multicolumn{1}{c}{25.96} & \multicolumn{1}{c}{61.06} & \multicolumn{1}{c}{88.30} & \multicolumn{1}{c}{30.73} & \multicolumn{1}{c}{59.50}\\
    \multicolumn{1}{l}{ICDPO} & \multicolumn{1}{c}{63.91} & \multicolumn{1}{c}{-23.27} & \multicolumn{1}{c}{20.31} & \multicolumn{1}{c}{91.56} & \multicolumn{1}{c}{16.33} & \multicolumn{1}{c}{53.94} & \multicolumn{1}{c}{85.10} & \multicolumn{1}{c}{21.23} & \multicolumn{1}{c}{53.16}\\
    \multicolumn{1}{l}{ICL} & \multicolumn{1}{c}{52.73} & \multicolumn{1}{c}{-32.05} & \multicolumn{1}{c}{10.33} & \multicolumn{1}{c}{88.00} & \multicolumn{1}{c}{4.74} & \multicolumn{1}{c}{46.36} & \multicolumn{1}{c}{75.46} & \multicolumn{1}{c}{16.38} & \multicolumn{1}{c}{45.91}\\
    \multicolumn{1}{l}{ICL$_\text{uni}$} & \multicolumn{1}{c}{50.85} & \multicolumn{1}{c}{-33.44} & \multicolumn{1}{c}{8.70} & \multicolumn{1}{c}{88.62} & \multicolumn{1}{c}{2.16} & \multicolumn{1}{c}{45.38} & \multicolumn{1}{c}{72.72} & \multicolumn{1}{c}{15.32} & \multicolumn{1}{c}{45.51}\\
    \bottomrule
  \end{tabular}
}
\caption{\label{ablation_results}
    Ablation study on \textit{HH-RLHF}. 
}
\end{table*}
\subsection{Main Results}
Automatic evaluations are conducted on both \textit{HH-RLHF} and \textit{SyntheticGPT}.
We deploy base models and their SFT variants on each dataset, utilizing LoRA~\cite{hu2021lora} to accommodate the limitations of constrained devices.
Since \modelname{} essentially borrows the capabilities of superior LLMs, we also deploy two \textit{borrowing} baselines, RM-BoN and RM-Aug, based on the Best-of-N policy and \citet{mudgal2023controlled}, respectively.
RM-BoN and RM-Aug can utilize the logits of superior LLMs as the external scorer~\cite{fu2023gptscore} to select the best response or intermediate block during decoding.
Although we introduce both LLaMA2-chat and GPT-3.5-turbo as the teachers, the detailed log probability of prompt tokens from GPT-3.5-turbo appears to be \textbf{inaccessible}, so we must compare \modelname{} and the two baselines using only LLaMA2-chat on \textit{HH-RLHF} and \textit{SyntheticGPT}.

As to \modelname{}, we evaluate its original version~(supported by randomly sampled demonstrations) and variants with only $\hat{S}$ or both $\hat{S}$ and retriever $R$. We accordingly set the following research questions~(RQs) to guide experiments:
\subsubsection*{RQ1: How does ICDPO perform well?}
Table~\ref{main_results1} presents the main results for \textit{HH-RLHF}, while those for \textit{SyntheticGPT} are provided in Appendix~\ref{appendix:main}.
Essentially, all methods show notable improvements over the corresponding base models. However, in the specific scenario where LLaMA2-chat is referenced, \modelname{} exhibits significant progress compared to RM-Aug and RM-BoN.
Overall, \modelname{} generally demonstrates competitive performance against SFT despite not undergoing fine-tuning.
These results strongly support the effectiveness of \modelname{}.

Furthermore, we observed that each method could receive lower scores in the domain of \textbf{Helpful} compared to \textbf{Harmless}.
We infer that \textbf{Helpful} needs more substantial content from base models or external sources, whereas \textbf{Harmless} may only require simpler stylistic changes.
Thus, Mistral, being the superior model combined with SFT where downstream information is forcibly integrated, achieves the highest scores in the \textbf{Helpful} domain.
However, \modelname{} also effectively enhances \textbf{Helpful} for Mistral, activated by contextual demonstrations, which is second only to SFT.
\subsubsection*{RQ2: How demonstrations affect ICDPO?} 
Intuitively, the quality of data, i.e. HPA degree, should heavily impact performance.
For instance, GPT-3.5-turbo can generally provide greater assistance for SFT with higher-quality samples compared to ordinary sources, as proved in \citet{song2023pro}.
\modelname{} hereby reflects similar trends.
According to Table~\ref{main_results2}, GPT-3.5-turbo and/or LLaMA2-chat can achieve higher scores than the original samples, consistent with Table~\ref{main_results1} where ICDPO demonstrates improvements from superior demonstrations.
This suggests that the meta-optimization in ICL does indeed function. In \S~\ref{sec:ablation}, we will provide a detailed analysis of the effects of $S$ using these higher-quality demonstrations.

Despite GPT-3.5-turbo being more powerful than LLaMA2-chat according to Table~\ref{main_results2}, \modelname{} seems better with demonstrations from LLaMA2-chat than GPT-3.5-turbo.
Believing it is not a coincidence, we make further analyses in Appendix~\ref{appendix:loss}.
\subsubsection*{RQ3: The impact of extra modules?} 
\modelname{} relies on $S$ and randomly sampled demonstrations by default.
In Table~\ref{main_results1}, we also test \modelname{} with only $\hat{S}$, or $\hat{S}+R$ which additionally involves the retriever $R$.
The overall performance can be improved step by step, except that $R$ with samples from the original datasets fails.
We attribute these results to the quality of the samples, as $R$ essentially narrows the gap between demonstrations and the tested sample.
Thus, if the initially chosen/rejected samples are not sufficiently \textit{good}/\textit{bad}, the estimation of $S$ collapses, and $R$ further exacerbates the confusion through meta-optimization.
\subsection{Ablation Study}
\label{sec:ablation}
In this section, we test the effectiveness of the remaining modules.
Our experiments focus on the variants of \textit{HH-RLHF} derived from LLaMA2-chat and GPT-3.5-turbo, as presented in Table~\ref{ablation_results}.\\
\textbf{Retriever $R$}\quad
We analyze the impact of fine-grained and coarse-grained retrieval with SBERT and BM25, respectively.
The results indicate that the latter approach~(\modelname{}$+\text{BM25}$ \textit{vs.} \modelname{}) can strongly enhance the meta-optimization in ICL, similar to genuine fine-tuning.
However, the former one~(\modelname{}$+R$ \textit{vs.} \modelname{}$+\text{BM25}$) occasionally results in marginal improvement (LLaMA2/Mistral on \textit{HH-RLHF}+GPT-3.5-turbo) or even a decline (Mistral on \textit{HH-RLHF}+LLaMA2-chat).
They occur upon powerful LLMs~(e.g. LLaMA2/Mistral against LLaMA) achieving high performance without SBERT, indicating that fine-grained retrieval provides greater benefits to weaker LLMs for strong LLMs can directly handle ICL well.
\\
\textbf{Contrastive Score $S$}\quad
Without $S$, \modelname{} degenerates into the normal ICL. 
We thus experiment with two decoding strategies: randomly selecting 1 from 3 candidates, and generating just 1 candidate\footnote{We also evaluate greedy search, which exhibits similar performance.}.
Obviously, ICL without selections from $S$ experiences significant performance declines, regardless of the decoding strategies. 
This validates the significance of $S$ as the key element in \modelname{}.
Since $S$ is a potential ranker, we also evaluate its performance in this aspect, as discussed in \S~\ref{sec:consistency}.
\begin{figure}[t]
    \centering
    \includegraphics[width=0.93\linewidth]{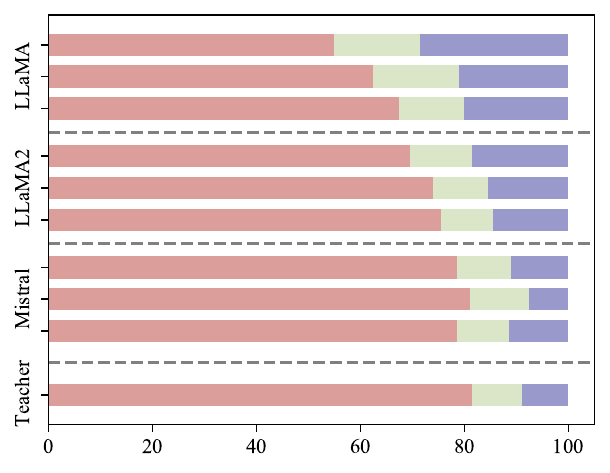} 
    \caption{GPT-4 computed win-rates of \modelname{} against golden responses in \textit{HH-RLHF}, using demonstrations from the teacher~(i.e. LLaMA2-chat). For each block titled by one base model, the bars from top to bottom are \modelname{}, \modelname{}$+\hat{S}$ and \modelname{}$+\hat{S}R$, while \textbf{red}, \textbf{light green} and \textbf{purple} represent the proportion of \textbf{win}, \textbf{tie} and \textbf{lose}, respectively.
    }
    \label{fig:gpt4}
\end{figure}
\subsection{GPT-4 Evaluation}
We implement GPT-4 evaluation as an additional validation of automatic evaluation with RM$_\text{test}$, following~\citet{song2023pro,liu2023aligning}.
We randomly select 200 samples from the test sets of \textit{HH-RLHF} and evaluate \modelname{}, \modelname{}$+\hat{S}$, \modelname{}$+\hat{S}R$, and their corresponding teachers. Their decoded responses are compared with the annotated choices in \textit{HH-RLHF}$_\text{raw}$ to compute the win rate.
In Figure~\ref{fig:gpt4}, we use demonstrations from LLaMA2-chat for \modelname{}, with LLaMA2-chat serving as the teacher model. The results for GPT-3.5-turbo can be found in Appendix~\ref{appendix:gpt4}.

Initially, we consider placing the tested candidates in the prompt from double directions to mitigate positional bias, as discussed in \citet{wang2023large}. However, several attempts yield similar results regardless of the direction.
We attribute it to the enhanced capabilities of GPT-4-32K and therefore use uni-directional tests to reduce costs.

We note that the results in Figure~\ref{fig:gpt4} align with those in Table~\ref{main_results1}, thereby validating the fairness of RM$_\text{test}$.
Generally, \modelname{} with $\hat{S}$ and $R$ outperforms \modelname{} without them.
With the more powerful base model, the third block~(Mistral) can even approach the performance of LLaMA2-chat.
\section{Discussion}
\begin{figure*}[t]
    \centering
    \includegraphics[width=0.95\textwidth]{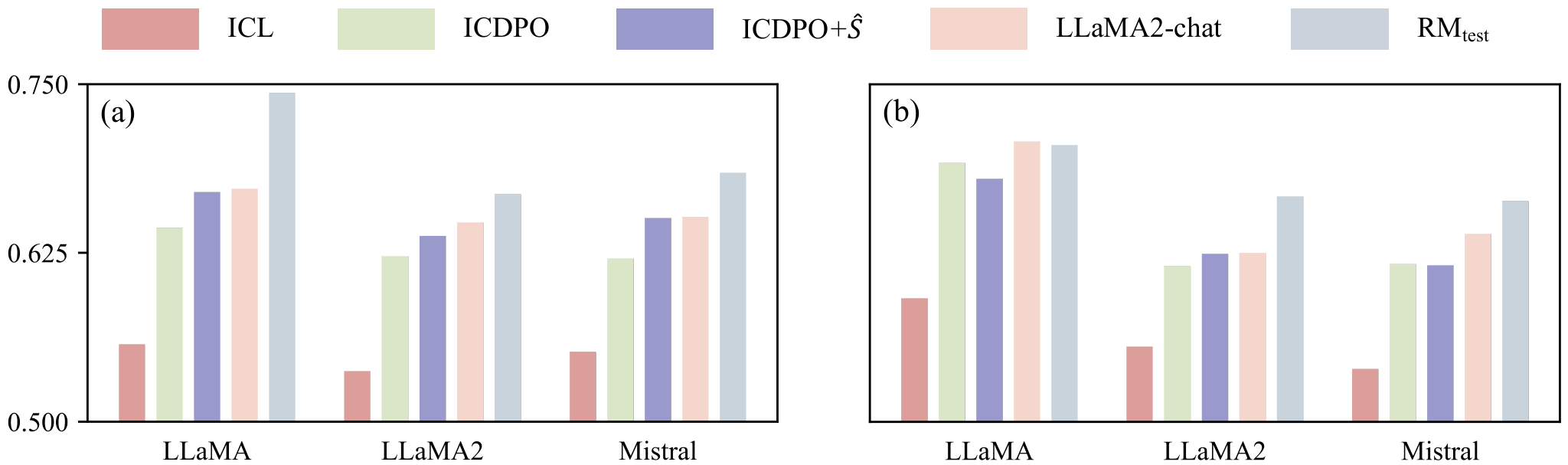} 
    \caption{Results of consistency between different scorers and GPT-4.
    We compute MRR to measure the degree of consistency. (a)~Results with randomly selected demonstrations. (b)~Results with demonstrations retrieved by $R$.}
    \label{fig:rm_alignment}
\end{figure*}

\subsection{Extension of Contrastive Score}
The contrastive score $S$ utilizes the optimized $\pi^*$ and initial $\pi_0$ to sort the candidates.
Since ICL can be one of the implementation methods for $\pi^*$, other methods should also be able to utilize $S$.

Consequently, we implement DPO + LoRA using the TRL package~\cite{vonwerra2022trl}, with $\pi$ defined as the $n$-th root of $P_{\pi}(y\mid x)$ to match the definition in DPO.
We evaluate the performance with and without $S$ (Table~\ref{dpo_results}), demonstrating that $S$ can still enhance DPO.
It indicates that $S$ may be a promising way for general use in HPA.
\begin{table}
\centering
\scalebox{0.86}{
  \begin{tabular}{llll}
    \toprule
    \multicolumn{1}{l}{\textbf{Method}} & \multicolumn{1}{c}{\textbf{LLaMA}} & \multicolumn{1}{c}{\textbf{LLaMA2}} & \multicolumn{1}{c}{\textbf{Mistral}}\\
    \midrule
    \multicolumn{4}{c}{w/ LLaMA2-chat}\\
    \midrule
    \multicolumn{1}{l}{SFT} & \multicolumn{1}{c}{34.73} & \multicolumn{1}{c}{57.59} & \multicolumn{1}{c}{63.43}\\
    \multicolumn{1}{l}{DPO} & \multicolumn{1}{c}{43.02} & \multicolumn{1}{c}{68.34} & \multicolumn{1}{c}{69.26}\\
    \multicolumn{1}{l}{DPO$+{S}$} & \multicolumn{1}{c}{48.11} & \multicolumn{1}{c}{71.62} & \multicolumn{1}{c}{71.84}\\
    \midrule
    \multicolumn{4}{c}{w/ GPT-3.5-turbo}\\
    \midrule
    \multicolumn{1}{l}{SFT} & \multicolumn{1}{c}{19.05} & \multicolumn{1}{c}{52.87} & \multicolumn{1}{c}{76.49}\\
    \multicolumn{1}{l}{DPO} & \multicolumn{1}{c}{30.88} & \multicolumn{1}{c}{95.00} & \multicolumn{1}{c}{86.61}\\
    \multicolumn{1}{l}{DPO$+{S}$} & \multicolumn{1}{c}{40.15} & \multicolumn{1}{c}{95.58} & \multicolumn{1}{c}{90.73}\\
    \bottomrule
  \end{tabular}
}
\caption{\label{dpo_results}
    DPO results on \textit{HH-RLHF}.
}
\end{table}
\subsection{Consistency of Scoring}
\label{sec:consistency}
\modelname{} computes the contrastive score $S$ to rank sampled candidates ${y}$ from ICL for the prompt $x$, similar to the methodology of RM$_{\text{test}}$. Therefore, we intend to evaluate \modelname{} as the ranking model.

We introduce \modelname{}, its enhanced version, \modelname{}$+\hat{S}$, and its simplified variant~(i.e., using only $\pi^*$ for scoring, denoted as ICL), alongside RM$_\text{test}$.
LLaMA2-chat is also incorporated as a reward model, like how it is used in RM-Aug and RM-BoN.
We set up two scenarios: one depicted in Figure~\ref{fig:rm_alignment}(a), where demonstrations for \modelname{} are randomly selected, and the other depicted in Figure~\ref{fig:rm_alignment}(b), which involves the proposed retriever $R$.
In each scenario, we select 200 samples, each containing 3 candidate responses sampled from the base model through ICL and sorted by GPT-4 as the ground truth.
We use the Mean Reciprocal Rank~(MRR) as the metric to fairly evaluate the competence of each method as a scorer and ranker.

Figure~\ref{fig:rm_alignment} illustrates that RM$_\text{test}$ achieves the highest performance in most cases, followed by LLaMA2-chat.
\modelname{} also performs well, with \modelname{}$+\hat{S}$ generally yielding equal or higher MRR scores, even approaching the performance of LLaMA2-chat as the teacher.
However, the performance of $\pi^*$ itself is unsatisfactory, significantly lagging behind others.
These findings exhibit that \modelname{} is a potent scorer beyond the vanilla ICL and approaches the performance of LLaMA2-chat through effective \textit{borrowing}.
\section{Related Work}
\subsection{Human Preference Alignment}
To mitigate the risk of generating toxic content, LLM should be aligned with human preference~\cite{wang2023aligning}, i.e. Human preference alignment~(HPA), which has been advanced through RLHF~\cite{ouyang2022training, zhu2024iterative, yu2023constructive,jang2023personalized, dai2023safe} and SFT methods~\cite{yuan2023rrhf, song2023pro, rafailov2023direct,wang2023making,zhang2023knowledgeable,liu2023statistical,xu2023reasons, hong2023cyclealign}.
DPO~\cite{rafailov2023direct} can be the representative one. 
It builds the relation between the RM and the combination of pre/post-optimized policies by transforming RLHF objective, which is inserted into reward modeling to derive an elegant SFT objective.

Nevertheless, fine-tuning LLMs is still costly. It triggers the need for fine-tuning-free methods, relying on self-selection~\cite{li2023rain}, external expert selection~\cite{mudgal2023controlled} or refinement of prompts~\cite{cheng2023black}. The proposed \modelname{} similarly refers to external experts, but does selection with self-estimation, which is based on reverse derivation of the relation in DPO.

\subsection{In-Context Learning}
LLM has the potential of instant few-shot learning through demonstrations in the context~\cite{brown2020language, chowdhery2023palm, dong2023survey, zheng-etal-2023-edit, yang-etal-2023-demonstration}, named In-Context Learning~(ICL). The underlying mechanism of ICL has also been carefully studied. From the perspective of information flow, \citet{wang-etal-2023-label} distinguish the different roles of upper and lower layers in LLMs for ICL, while \citet{dai-etal-2023-gpt} established a dual relation between gradient descent and Transformer attention, thus illustrating that ICL as a meta-optimizer can be similar to explicit fine-tuning. We extend it to HPA, where the optimized policy can be easily acquired for generation and scoring without fine-tuning.
\section{Conclusion}
In this paper, we equip LLMs with HPA by leveraging capabilities from superior models without the need for costly fine-tuning.
We rethink the procedures of DPO and focus on the crucial relation between the RM and the optimized policy.
Building upon this relation, we propose \modelname{}.
It implements ICL to instantly optimize the LLM, which through collaboration with the initial policy can effectively estimate the degree of HPA and enhance the final performance.
Comprehensive experiments demonstrate the effectiveness of \modelname{} across various forms, encompassing both content generation and scoring.
We hope this work to be a catalyst for further exploration of fine-tuning-free methods towards HPA.

\section{Ethics Statement}
We observe that the data involved in this work may indispensably contain sensitive, offensive, and misleading content, whose presence does not represent our attitudes, but is solely for research and should not be used or distributed outside of research contexts.

We are committed to establishing a more inclusive and ethically sound era of AI technology, which can be applied to legitimate needs and generate content that aligns with universally positive human values. 

\section{Limitations}
\modelname{} has been shown powerful but user-friendly, because it is fine-tuning-free and learns effectively from just demonstrations from superior LLMs. Although we conduct abundant experiments to evaluate \modelname{} comprehensively, there remain a few aspects of limitation:
\\
1. Despite 7B LLMs showing the satisfying capability of ICL, we fail to evaluate \modelname{} on larger models for their costly requirements on hardware.
\\
2. Similarly, we do not test the effect of changes in the number of demonstrations for ICL. Nonetheless, we believe it should further boost \modelname{} with increasing demonstrations.
\\
Due to limited computational resources, we leave them to the community with interest for further exploration.

\bibliography{anthology,custom}

\appendix
\section{Dataset Preparation}
\label{appendix:data}
We introduce the following two datasets for \modelname{}:
\begin{itemize}
    \item \textit{HH-RLHF} is proposed by \citet{bai2022training}, focusing on the domain of harmlessness and helpfulness in multi-turn conversations. While it initially consists of four subsets, we select two representative ones: \textit{harmless-base} and \textit{helpful-base}, which we denote as \textbf{Harmless} and \textbf{Helpful}, respectively. We mix the data of two domains for training, while separately evaluating each method in the main experiment.
    \item \textit{SyntheticGPT}\footnote{\url{https://huggingface.co/datasets/Dahoas/synthetic-instruct-gptj-pairwise}} collects about 33.1K samples of instruction following. Since its original version just has a training set, we manually split it into train/dev/test ones.
\end{itemize}
Each sample in these datasets has two candidates, including a shared prompt and two chosen/rejected candidate responses. In order to alleviate the pressure of GPU memory and accelerate the inference, we filter all samples according to sequence length in advance, 320/128 tokens for prompts/responses in \textit{HH-RLHF}, while 128/200 in \textit{SyntheticGPT}.

\section{Implementation Details}
\label{appendix:implementation}
We implement \modelname{} with all base models on Huggingface.Library~\cite{wolf-etal-2020-transformers}. For ICL, the number of demonstrations and top-p sampling is 2 and 3, respectively, where p is set to 0.8.
To facilitate demonstration retrieval in ICL, we deploy BM25 and SBERT\footnote{\url{https://huggingface.co/sentence-transformers/all-mpnet-base-v2}}. The BM25 model first retrieves 20 samples, which are then re-ranked by the SBERT retriever to obtain highly semantically similar ones.
The templates for ICL have been placed in Appendix~\ref{appendix:template} for a detailed overview.

Furthermore, the third-party reward model for automatic scoring is denoted as RM$_{\text{test}}$\footnote{\url{https://huggingface.co/OpenAssistant/oasst-rm-2-pythia-6.9b-epoch-1}}, while GPT-4-32K is employed for GPT-4 evaluation. To carry out \textit{borrowing} in ICL, we employ LLaMA2-chat to generate new choices for \textit{HH-RLHF} and \textit{SyntheticGPT}, while for GPT-3.5-turbo, we reuse \textit{HH-RLHF}$_{\text{ChatGPT}, 3}$ released by \citet{song2023pro}. The whole details can be found in the released code.

\section{Additional Main Results}
\label{appendix:main}
\begin{table}[h]
\centering
\scalebox{0.8}{
  \begin{tabular}{llll}
    \toprule
    \multicolumn{1}{l}{\textbf{Method}} & \multicolumn{1}{c}{\textbf{LLaMA}} & \multicolumn{1}{c}{\textbf{LLaMA2}} & \multicolumn{1}{c}{\textbf{Mistral}}\\
    \midrule
    \multicolumn{4}{c}{w/ \textit{SytheticGPT}$_\text{raw}$}\\
    \midrule
    \multicolumn{1}{l}{Base} & \multicolumn{1}{c}{-121.85} & \multicolumn{1}{c}{-101.01} & \multicolumn{1}{c}{34.75}\\
    \multicolumn{1}{l}{SFT} & \multicolumn{1}{c}{36.05} & \multicolumn{1}{c}{77.04} & \multicolumn{1}{c}{96.10}\\
    \multicolumn{1}{l}{ICDPO} & \multicolumn{1}{c}{27.74} & \multicolumn{1}{c}{77.89} & \multicolumn{1}{c}{68.97}\\
    \multicolumn{1}{l}{ICDPO$+{\hat{S}}$} & \multicolumn{1}{c}{29.37} & \multicolumn{1}{c}{83.34} & \multicolumn{1}{c}{78.34}\\
    \multicolumn{1}{l}{ICDPO$+{\hat{S}}R$} & \multicolumn{1}{c}{53.97} & \multicolumn{1}{c}{72.26} & \multicolumn{1}{c}{71.69}\\
    \midrule
    \multicolumn{4}{c}{w/ LLaMA2-chat}\\
    \midrule
    \multicolumn{1}{l}{SFT} & \multicolumn{1}{c}{41.89} & \multicolumn{1}{c}{99.21} & \multicolumn{1}{c}{99.09}\\
    \multicolumn{1}{l}{RM-Aug} & \multicolumn{1}{c}{-95.63} & \multicolumn{1}{c}{-77.14} & \multicolumn{1}{c}{84.48}\\
    \multicolumn{1}{l}{RM-BoN} & \multicolumn{1}{c}{-97.38} & \multicolumn{1}{c}{-70.08} & \multicolumn{1}{c}{88.00}\\
    \multicolumn{1}{l}{ICDPO} & \multicolumn{1}{c}{49.82} & \multicolumn{1}{c}{100.41} & \multicolumn{1}{c}{113.67}\\
    \multicolumn{1}{l}{ICDPO$+{\hat{S}}$} & \multicolumn{1}{c}{50.36} & \multicolumn{1}{c}{102.09} & \multicolumn{1}{c}{118.37}\\
    \multicolumn{1}{l}{ICDPO$+{\hat{S}}R$} & \multicolumn{1}{c}{96.82} & \multicolumn{1}{c}{111.39} & \multicolumn{1}{c}{119.10}\\
    \bottomrule
  \end{tabular}
}
\caption{\label{main_results3}
    Main results on \textit{SyntheticGPT}.
}
\end{table}

\section{Additional Results of GPT-4 Evaluation}
\label{appendix:gpt4}
\begin{figure}[htbp]
    \centering
    \includegraphics[width=0.93\linewidth]{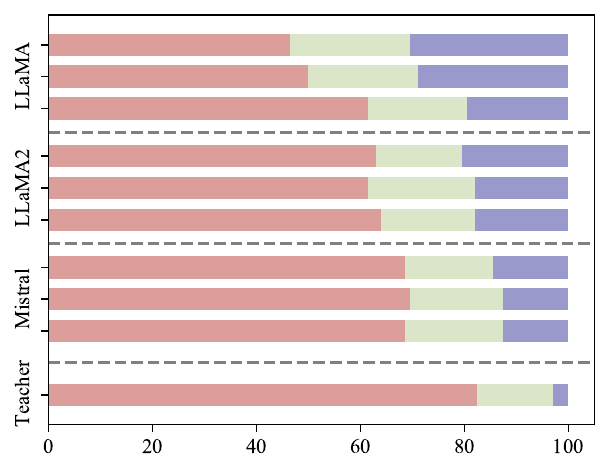} 
    \caption{GPT-4 computed win-rates of \modelname{} against golden responses in \textit{HH-RLHF}, using demonstrations from the teacher~(i.e. GPT-3.5-turbo). For each block titled by one base model, the bars from top to bottom are \modelname{}, \modelname{}$+\hat{S}$ and \modelname{}$+\hat{S}R$, while \textbf{red}, \textbf{light green} and \textbf{purple} represent the proportion of \textbf{win}, \textbf{tie} and \textbf{lose}, respectively.
    }
    \label{fig:gpt4_2}
\end{figure}

\section{Distribution of Demonstrations}
\label{appendix:loss}
\begin{figure}[htbp]
    \centering
    \includegraphics[width=\linewidth]{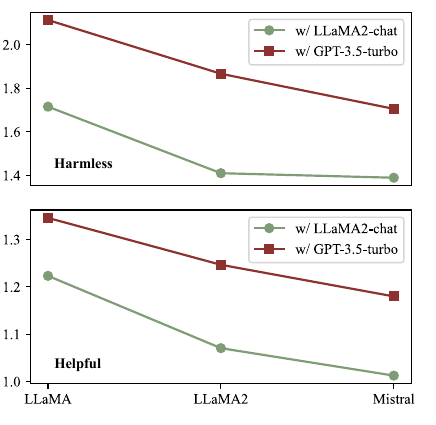} 
    \caption{Loss of different base models on demonstrations from LLaMA2-chat and GPT-3.5-turbo.}
    \label{fig:loss}
\end{figure}
Although GPT-3.5-turbo surpasses LLaMA2-chat in Table~\ref{main_results2}, utilizing demonstrations from LLaMA2-chat leads to better performance of \modelname{}. 
Since ICL can be regarded as an instant LLM fine-tuning, we speculate that responses from LLaMA2-chat can be closer to the distribution of open-source LLMs, like LLaMA, than those from GPT-3.5-turbo, which mitigates the difficulty of ICL on these samples. Therefore, this should be illustrated by computing the NLL loss on demonstrations of both sources, where a smaller value suggests a closer distribution.

We hereby compute the loss with \textbf{mean} rather than \textbf{sum} reduction to eliminate the impact of sequence length on the magnitude of values, as depicted in Figure~\ref{fig:loss}. All 3 base models exhibit significantly smaller losses on demonstrations from LLaMA2-chat, thus verifying the hypothesis above. The wider gap in distribution may arise because GPT-3.5-turbo mainly relies on private data, resulting in distinctions in style or other aspects compared to open-source LLMs based on public data.

\section{Prompt Templates for ICL}
\label{appendix:template}
Templates for $\pi(y \mid [\textbf{d}^+; x])$ and $\pi(y \mid [\textbf{d}^-; x])$ are illustrated as Figure~\ref{pos_prompt} and \ref{neg_prompt}, respectively.

\begin{figure*}[t]
  \centering
  \begin{tcolorbox}
    \#You are asked to generate a response for each given dialogue context. The last response is generated by you, which should be as **polite/harmless/helpful/honest** as possible:\\
    
    \#\#

    [The Context of No.1 demonstration]
    
    \#\#\#Generate a **polite/harmless/helpful/honest** response: [The positive response of No.1 demonstration]\\

    \#\#

    [The Context of No.2 demonstration]
    
    \#\#\#Generate a **polite/harmless/helpful/honest** response: [The positive response of No.2 demonstration]\\

    \#\#

    [The Context of tested sample]
    
    \#\#\#Generate a **polite/harmless/helpful/honest** response:
  \end{tcolorbox}
  \caption{\label{pos_prompt}
  The prompt template used to trigger LLMs generating preferred content.}
\end{figure*}

\begin{figure*}[t]
  \centering
  \begin{tcolorbox}
    \#You are asked to generate a response for each given dialogue context. The last response is generated by you, which should be as **offensive/harmful/helpless/misleading** as possible:\\
    
    \#\#

    [The Context of No.1 demonstration]
    
    \#\#\#Generate an **offensive/harmful/helpless/misleading** response: [The negative response of No.1 demonstration]\\

    \#\#

    [The Context of No.2 demonstration]
    
    \#\#\#Generate an **offensive/harmful/helpless/misleading** response: [The negative response of No.2 demonstration]\\

    \#\#

    [The Context of tested sample]
    
    \#\#\#Generate an **offensive/harmful/helpless/misleading** response:
  \end{tcolorbox}
  \caption{\label{neg_prompt}
  The prompt template used to trigger LLMs generating non-preferred content.}
\end{figure*}

\end{document}